\documentclass[journal]{IEEEtran}
\usepackage[utf8]{inputenc}
\DeclareUnicodeCharacter{2194}{-}
\usepackage{times}
\usepackage{epsfig}
\usepackage{graphicx}
\usepackage{amsmath}
\usepackage{amssymb}
\usepackage{multirow}
\usepackage{cite}
\usepackage{booktabs}
\usepackage{balance}
\usepackage{color}
\usepackage{float}
\usepackage[pagebackref=true,breaklinks=true,colorlinks,bookmarks=false]{hyperref}

\usepackage[dvipsnames,svgnames,x11names]{xcolor}

\newcommand{\eg}{\emph{e.g.}}
\newcommand{\ie}{\emph{i.e.}}

\begin{document}


\title{SPG-VTON: Semantic Prediction Guidance \\ for Multi-pose Virtual Try-on}

\author{Bingwen~Hu,
        Ping~Liu,~\IEEEmembership{Member,~IEEE,}
        Zhedong~Zheng,
        and Mingwu~Ren

\thanks{Manuscript received April 19, 2005; revised August 26, 2015.}}%
\markboth{Journal of \LaTeX\ Class Files,~Vol.~14, No.~8, August~2015}%
{Shell \MakeLowercase{\textit{et al.}}: Bare Demo of IEEEtran.cls for IEEE Journals}


\maketitle


\begin{abstract}
Image-based virtual try-on is challenging in fitting target in-shop clothes onto a reference person under diverse human poses. Previous works focus on preserving clothing details (\emph{e.g.,} texture, logos, patterns) when transferring desired clothes onto a target person under a fixed pose. However, the performances of existing methods significantly dropped when extending existing methods to multi-pose virtual try-on. In this paper, we propose an end-to-end semantic prediction guidance multi-pose virtual try-on network (SPG-VTON), which can fit the desired clothing onto a reference person under arbitrary poses. Specifically, SPG-VTON is composed of three submodules. First, a semantic prediction module (SPM) generates the desired semantic map. The predicted semantic map provides more abundant guidance to locate the desired clothing region and produce a coarse try-on image. Second, a clothes warping module (CWM) warps in-shop clothes to the desired shape according to the predicted semantic map and the desired pose. Specifically, we introduce a conductible cycle consistency loss to alleviate the misalignment in the clothing warping process. Third, a try-on synthesis module (TSM) combines the coarse result and the warped clothes to generate the final virtual try-on image, preserving details of the desired clothes and under the desired pose. In addition, we introduce a face identity loss to refine the facial appearance and maintain the identity of the final virtual try-on result at the same time. We evaluate the proposed method on the most massive multi-pose dataset (MPV) and the DeepFashion dataset. The qualitative and quantitative experiments show that SPG-VTON is superior to the state-of-the-art methods and is robust to data noise, including background and accessory changes, \ie, hats and handbags, showing good scalability to the real-world scenario.
\end{abstract}

\begin{IEEEkeywords}
Virtual Try-on, Multi-pose, Semantic Prediction, End-to-end
\end{IEEEkeywords}

\IEEEpeerreviewmaketitle

\section{Introduction}
\begin{figure}[t]
		\begin{center}
            \centering
            \includegraphics[width=\linewidth]{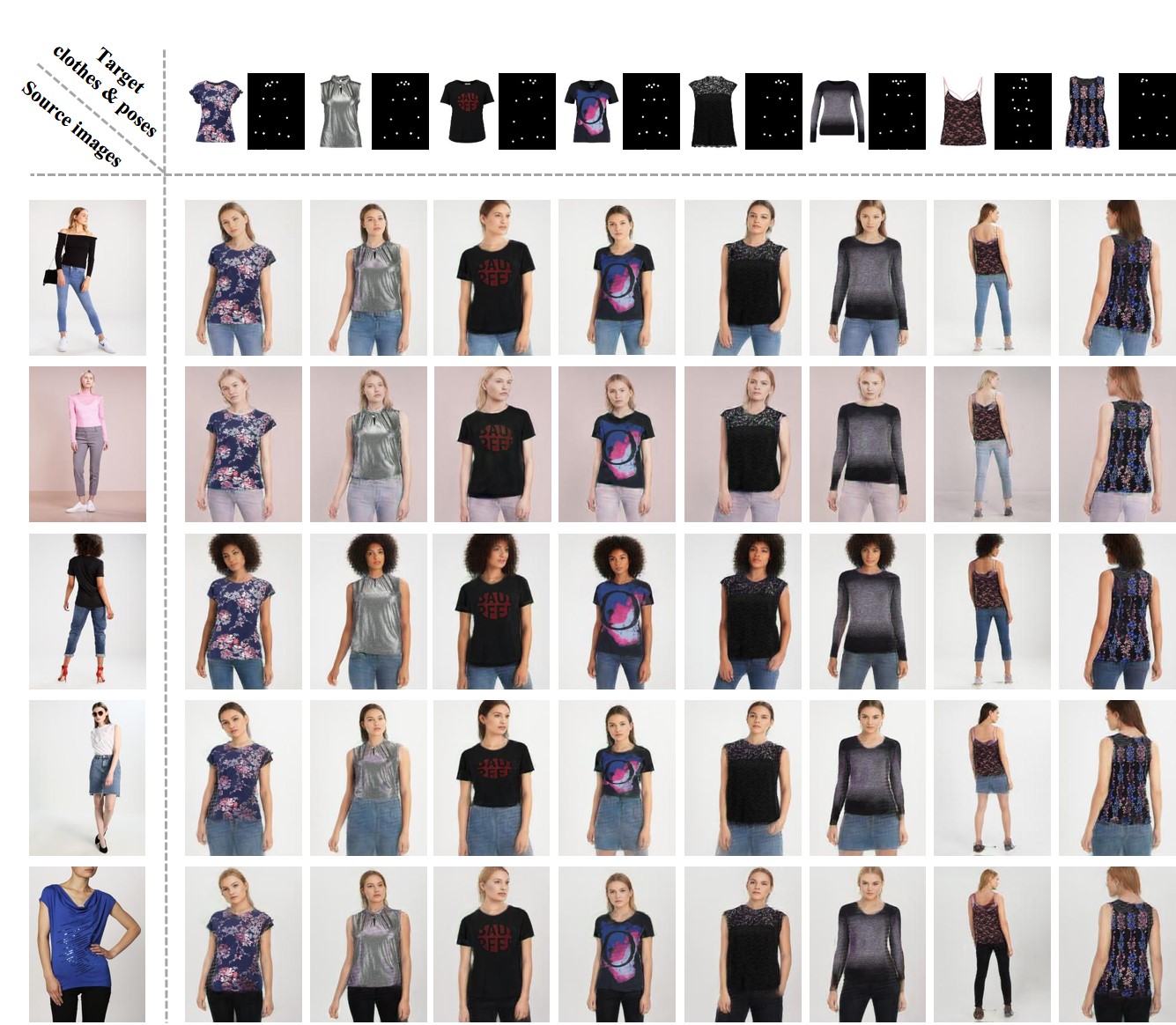}
        \end{center}
        \vspace{-.5em}
        \caption{ Visual results of multi-pose virtual try-on by the proposed method. First column: source images. Top row: in-shop clothes and target poses. Other columns: generated images. SPG-VTON produces photo-realistic virtual try-on results, which preserves both clothes details and person characteristics. Please zoom in to see the details of the generated images.
        } 
        \label{fig:multi-pose}
\end{figure}

Image-based virtual try-on systems aim at fitting target in-shop clothes onto a reference person, which is a branch of the field of image synthesis. Driven by the rapid development of image synthesis \cite{isola2017image,zhu2017unpaired,wang2018high,karras2019style}, the topic of image-based virtual try-on has attracted more interest and has vast potential applications in virtual reality and human-computer interaction. Despite significant progress in previous works~\cite{han2018viton,wang2018toward,dong2019towards,yu2019vtnfp,han2019clothflow,issenhuth2019end,zheng2019virtually,HanYang2020,chou2021template,ge2021parser,gao2021shape,Xie2021} for virtual try-on, multi-pose virtual try-on has not been well studied. Concretely, for a given person image, the virtual fitting system could generate realistic images of this person in different poses while preserving the desired clothes' appearance. The multi-pose virtual fitting system is more in line with practical application scenarios. The existing works on multi-pose virtual fitting tasks are insufficient, and there are problems such as mismatch between the target clothes and the given pose, distortion of the clothes region in the try-on result, and loss of details that need to be further explored. To solve these problems, we propose a method to build a robust multi-pose virtual try-on system based on 2D images (as shown in Fig.~\ref{fig:multi-pose}).



Most of the previous works~\cite{han2018viton,wang2018toward,yu2019vtnfp,han2019clothflow,zheng2019virtually,HanYang2020} focus on swapping clothes in a fixed pose without considering the body's changing posture. However, in realistic scenarios, users would like to intuitively see the results of wearing the given clothes in different poses. Recently, 
MG-VTON
~\cite{dong2019towards} made the first attempt at a virtual fitting system guided by multiple poses. MG-VTON is a multistage framework that includes a human parsing network, a warping generative adversarial network, and a refinement render network. Although MG-VTON makes a significant stride in virtual fitting system construction under arbitrary posture, it still has some limitations that restricts its further applications: (1) An end-to-end mechanism is missing in the MG-VTON training process. In MG-VTON, modules designed with different purposes are utilized to generate the desired clothing deformation under different postures, a coarse result, and a refined result step by step. Each module/step is optimized independently and cannot collaborate with others bidirectionally, which might cause a suboptimal result. (2) In the inference process, MG-VTON requires multiple steps to generate the final results, which might cost more human interventions and cause error accumulations. (3) MG-VTON only focuses on manipulation of body parts while ignoring the other parts, for example, faces. The absence of a mechanism to process nonbody parts makes their generated results unpleasant in nature and appearance.


To address the limitations mentioned above, we present a new multi-pose virtual try-on network that can work in an end-to-end manner. To make our method able to handle variations introduced by arbitrary postures, we propose to introduce semantic prior knowledge into our network, making the learning process receive additional guidance from semantic prior knowledge. We name our designed method semantic prediction guidance-virtual try on system, \textit{a.k.a.}, SPG-VTON. As shown in Fig. \ref{fig:overview}, the SPG-VTON consists of three major modules, including the semantic prediction module (SPM), the clothes warping module (CWM), and the try-on synthesis module (TSM). The purpose of the designed SPM is to predict the semantic map for target images, which is utilized to provide additional spatial and semantic guidance during the learning process. Given the semantic map for source images, the in-shop clothes, the target pose, and the predicted semantic map for the target image, a coarse result and corresponding predicted clothing mask are generated by SPM. The CWM is introduced to warp the in-shop clothes to the desired shape according to the semantic map predicted by the SPM. To alleviate the misalignment between the desired in-shop clothes and the target human posture, we propose utilizing a conductible cycle consistency constraint in our network learning. Given the target pose, the coarse result generated by SPM, and the warped clothes generated by CWM, TSM generates the final try-on image with high precision and realism. Furthermore, to make the generated results photo-realistic, we introduce a global discriminator and a local discriminator to control the global shape and local texture of the generated results; to make the generated image visually pleasant, we utilize a face identity constraint to keep the synthesis face region realistic.

Extensive experiments on the MPV dataset~\cite{dong2019towards} show that our method achieves superior performance to several existing approaches~\cite{han2018viton,wang2018toward,dong2019towards}. It is worth noting that what we studied in this work is currently the largest dataset for the multi-pose virtual try-on task. Since all images are collected from the internet, the dataset inevitably contains unexpected "label noise", such as misalignment and different backgrounds. The noise existing in web collected data inevitably compromises the training process in the dataset, making it challenging to train a robust virtual try-on system based on these noisy images. 
The proposed method is robust to noisy data through the mutual cooperation of the introduced various losses and the end-to-end model frameworks. The specific mechanism can be summarized as follows:
(1) In the training process, the end-to-end approach allows the modules to be integrated, dynamically adjusts the network parameters of each module, and encourages the model to generate better results. Meanwhile, the end-to-end generation could, in turn, correct inaccurate predicted semantic maps to guide the model with more accurate semantic information.
(2) This paper introduces global-oriented losses, \ie, reconstruction loss and perceptual loss, to make the generated result consistent with the ground-truth image at both the pixel level and perceptual level to resist the interference of noisy data. In addition, this paper also introduces locally oriented losses, \ie, the conductible cycle consistency loss and the face identity loss, to ensure that the clothing area and facial area of the generated image retain more characteristic information of face regions. Moreover, introducing global and local adversarial loss can also ensure that the generated image is close to the real image and prevent the negative impacts from noisy input.
(3) One way of compression or denoising in prior works~\cite{lu2015noise,qi2018structure,zhou2020image} is to extract kernel information through latent representation learning. The latent representation is usually much smaller in dimension than the original input data, making it easier to control and analyze. In this work, the role of prior semantic information (human semantic map) is similar to that of latent representation. Specifically, the first process of SPM predicts the semantic map of the target image, and the second process of SPM predicts the mask of the target clothing area. In this case, SPG-VTON can accurately locate the target clothing area by combining the semantic map of the target image and the target clothing mask, which could prevent the generation of background noise.
Benefiting from our designed method and exploration, we experimentally observe that the proposed method is still robust to such training noise and demonstrates good scalability to unseen test images during inference.
The main contributions of the proposed method are summarized as follows:
\begin{itemize}
\item We propose an end-to-end image-based multi-pose virtual try-on system called SPG-VTON, which can synthesize high-quality try-on images. Compared to previous works, the proposed method could fit the desired clothing onto a reference person under arbitrary poses while preserving details of the desired clothes.
\item We conduct extensive explorations and locate effective strategies for learning a robust and accurate virtual try-on network for multi-pose inputs. The novelty points of the proposed method can be summarized as follows:
(1) We introduce a conductible cyclic consistency loss to alleviate the misalignment in the clothing warping process. Concretely, the conductible cycle consistency loss could match the shape of the deformed desired clothes with the target person image and maintain the characteristics of the desired clothes in the generated try-on image.
(2) We introduce both global and local adversarial losses and face identity loss to refine the facial appearance and maintain the identity of the final virtual try-on result at the same time. Specifically, the role of global and local adversarial losses encourages the generated image to be close to the real image, whether it is the whole image or part of the whole image. Moreover, the role of face identity loss enforces that the identity of the generated image remains unchanged.
(3) We apply an end-to-end training strategy to boost the proposed method to generate accurate semantic maps and improve the virtual try-on results under pose transfer. Additionally, the end-to-end manner can effectively reduce human interventions and avoid error accumulations in the inference process.
\item The qualitative and quantitative experiments on two prevailing datasets, \ie, MPV~\cite{dong2019towards} and DeepFashion~\cite{liu2016deepfashion}, demonstrate the advantages of our method in virtual try-on, especially when given different postures with heavy variations. Ablation studies also show that the proposed method has good scalability to unseen test data and is robust to label noise in the training set. 
\end{itemize}

\section{Related Work}

\subsection{Pose-Guided Person Image Generation}
Pose-guided person image generation is a practical yet challenging topic. In past years, generative adversarial networks (GANs)~\cite{goodfellow2014generative} and various extensions~\cite{mirza2014conditional,isola2017image,zhu2017unpaired,zheng2017unlabeled, wang2018high,dong2019fashion, karras2019style,liu2019stgan,huang2020real} have made significant progress in this research direction. However, due to the high variations existing in human poses and human appearance, these previous works~\cite{mirza2014conditional,isola2017image,zhu2017unpaired, wang2018high,dong2019fashion, karras2019style,liu2019stgan} still suffer from their limited scalability in pose-guided person image generation. To generate high-quality person images in arbitrary poses, Ma~\emph{et al.}~\cite{ma2017pose} proposed a two-stage generation framework. ~\cite{ma2017pose} first uses the U-Net-like network to produce initial images with blur and then applies the adversarial method to refine coarse results. To further improve generated image qualities, Ma~\emph{et al.} proposes a~\textit{disentangling} strategy~\cite{ma2018disentangled}, encoding a given person image into three factors: pose, foreground, and background, which are decoded back to an image space after editing a specific factor. It is believed that manipulating those disentangled factors rather than treating them as a whole can benefit generation quality improvement. Zheng~\emph{et al.} disentangles the input pedestrian images into structure and appearance embedding and can easily exchange codes to generate a source person with target clothes~\cite{zheng2019joint}. \cite{esser2018variational} also adopts the disentanglement strategy. Specifically, they use a conditional U-Net architecture that combines the appearance decomposed from a variational autoencoder~\cite{kingma2013auto} with a given shape to reconstruct a new image. Methods such as~\cite{ma2017pose, ma2018disentangled,esser2018variational} focus on the~\textit{global} pose deformation between the source image and the target image while ignoring the~\textit{local} structure of generated images. Therefore, these methods have difficulty maintaining the local details of the original image, especially when there is a large pose discrepancy between the source image and the target image.
Recently, some works~\cite{grigorev2018coordinate,siarohin2018deformable,dong2018soft,song2019unsupervised, han2019clothflow} have focused on the spatial deformation relationship in pose changes. \cite{grigorev2018coordinate} uses an inpainting network to estimate the coordinates in source images for elements of the body surface. Def-GAN~\cite{siarohin2018deformable} designs deformable skip connections in the generator to address the pixel-to-pixel misalignment caused by the pose differences. \cite{dong2018soft,song2019unsupervised, han2019clothflow} 
employ a specific module to predict the human semantic map after the pose changes to align the source image with the target pose to enforce the module to generate high-quality images. In this work, we also introduce a semantic map prediction module to produce the semantic map under a given pose. The difference is that when our method predicts the semantic map under a given pose, the clothing region of the semantic map changes with the given clothes. In contrast, the works mentioned above for pose-guided person image generation do not involve this aspect.

\subsection{Virtual Try-on}
The image-based virtual try-on task is a particular case of person generation. The core difference is that this task aims to generate a person image while the clothes region is changed to the desired clothes.
The thin-plate spline~(TPS) \cite{bookstein1989principal} transformation is a typical 2-D interpolation model that performs geometric deformation between images by controlling a set of registration points between two images. VITON~\cite{han2018viton} directly applies shape context-based matching\cite{Belongie2002shape} to estimate TPS transformation parameters between the mask of desired clothes and the clothes mask of the target person. Furthermore, CP-VTON~\cite{wang2018toward} uses a learnable method to estimate the TPS transformation parameters via convolutional neural networks dynamically. 
In the image-based virtual try-on task, using convolutional neural networks to learn the TPS transformation parameters between the desired clothes and the given human image is verified as a practical approach. The pioneering works~\cite{ wang2018toward,yu2019vtnfp,issenhuth2019end,dong2019towards,zheng2019virtually,HanYang2020} mainly use two ways to estimate TPS transformation parameters. One way~\cite{wang2018toward,yu2019vtnfp,issenhuth2019end},
is to use the geometric matching network~\cite{rocco2017convolutional} to estimate the TPS transformation parameters between the target clothing and the person representation (embedding the body shape, the target pose and reserved regions of the source image). In addition, these methods directly use the pixelwise $\mathcal{L}_{1}$-norm between warped clothes and the clothing area extracted from the target image to train the geometric matching network. Another way~\cite{dong2019towards,zheng2019virtually,HanYang2020}
is to apply the geometric matching network to estimate the TPS transformation parameters between the clothing or clothing mask and the clothing area mask or the body shape obtained from the predicted semantic map.
Similar to the first way, these methods
also use the pixelwise $\mathcal{L}_{1}$-norm between warped clothes and the clothes region extracted from the target person image to train the geometric matching network. 

Although the methods mentioned above can produce high-quality fitting images to a certain extent, there is still a large gap between the generated images and natural images. For example, VITON~\cite{han2018viton} and CP-VTON~\cite{wang2018toward} are state-of-the-art virtual try-on approaches that adopt a multistage coarse-to-fine strategy to tackle the virtual try-on task of a single pose. However, neither of these two methods includes changes in human pose. In this case, these methods cannot avoid distortion and misalignment in the process of clothing deformation (such as the distortion and misalignment of texture, patterns, logos, and embroidery). 

\begin{figure*}[t]
  \centering
  \includegraphics[width=\linewidth]{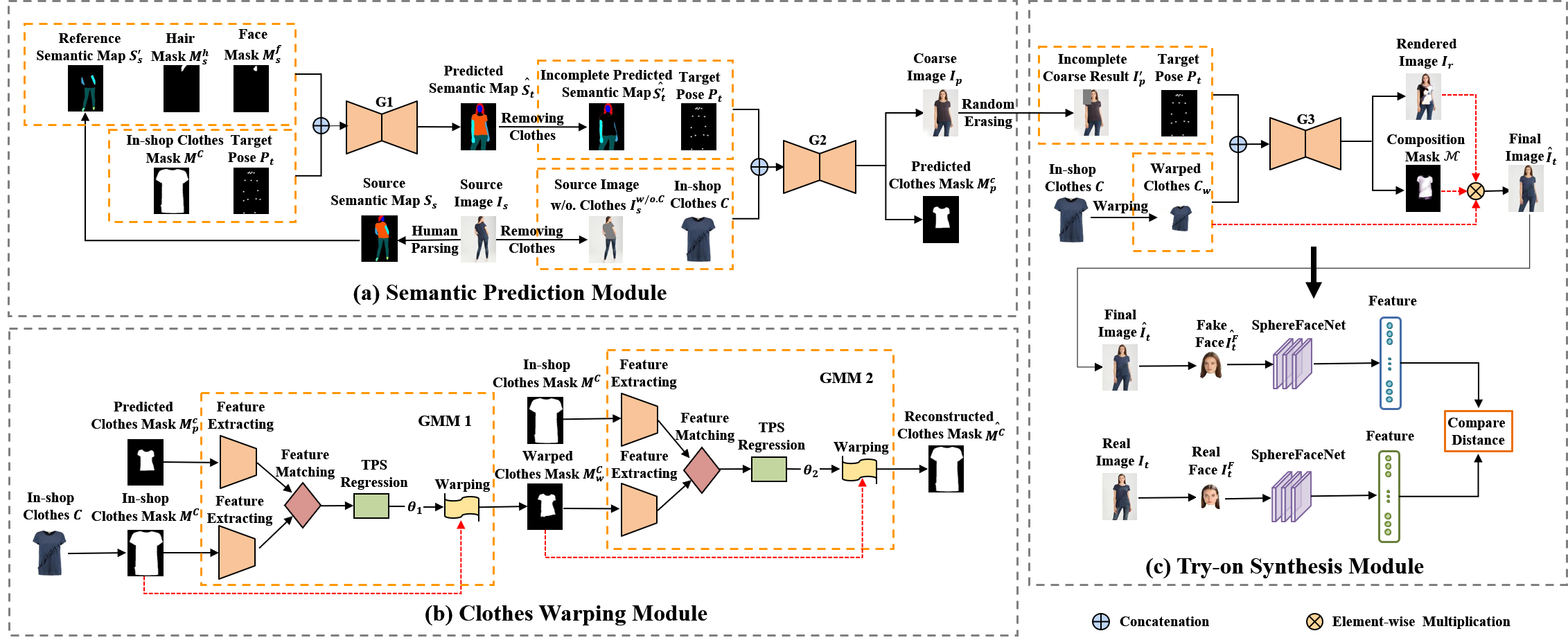}
\vspace{-2em}
  \caption{The overview of our SPG-VTON.  (a)~The Semantic Prediction Module (SPM) consists of two processes. One is the target semantic map prediction, and the other one is the target clothes mask prediction and coarse result generation. (b) The Clothes Warping Module (CWM) warps the in-shop clothes to the shape of the clothes region of the target image, according to the Thin-Plate Spline~(TPS) transformation parameters estimated between the mask of in-shop clothes and the predicted clothes mask. The CWM is composed of two Geometric Matching Modules (GMM)~\cite{rocco2017convolutional}.  In specific, the GMM 1 deforms the mask of the in-shop clothes to the same shape with the predicted clothes mask, and then the GMM 2 is used to convert the warped clothes mask back to the mask of the desired clothes. Note that we do not directly apply the $\mathcal{L}_{1}$-norm between the warped clothes mask and the predicted clothes, and $\mathcal{L}_{1}$-norm between the reconstructed clothes and the original clothes mask to train GMM 1 and GMM 2. By contrary, we introduce a conductible cycle consistency loss to indirectly constrain GMM 1 and GMM 2, respectively (see~\ref{3.3} for details). (c) The Try-on Synthesis Module (TSM) combines the incomplete coarse results, the target pose, and the warped desired clothes to synthesis the final virtual try-on images. Also, the pre-trained SphereFaceNet~\cite{liu2017sphereface} is applied to compare the distance between the generated face region and the ground-truth face region, which enforces the generator $G_{3}$ to generate realistic and natural faces.
  } 
  \label{fig:overview}
\end{figure*}

\section{Method}
We propose a novel image-based virtual try-on network named SPG-VTON, which focuses on multi-pose virtual try-on. Specifically, for a given source person image, a target in-shop clothes, and a target pose, the proposed method aims to generate a new person image such that the same person wears the target in-shop clothes and preserves the target pose. That is, given different poses, the proposed method can generate high-quality virtual try-on images. 

The SPG-VTON consists of three submodules, including the semantic prediction module (Section~\ref{3.2}), the clothes warping module (Section~\ref{3.3}), and the try-on synthesis module (Section~\ref{3.4}). We show the overview of SPG-VTON in Fig.~\ref{fig:overview}. Concretely, the SPM has two sequential processes. The first process of SPM aims to predict the semantic map of the target image according to the source semantic map, the in-shop clothes, and the target pose. The predicted semantic map provides precise guidance to locate the region of the desired clothes and generate a coarse virtual try-on image. Then, we combine the predicted semantic map, the in-shop clothes, and the target pose as the input of the second process of SPM to generate the coarse result and the predicted clothing mask. Subsequently, the CWM warps the in-shop clothes to the desired shape according to the predicted semantic map.

\subsection{Person Representation}
The diverse clothes and human poses struggle with the performance of the virtual try-on system. During the training process, the human pose and human body semantic map are critical supervision information for understanding the human geometric structure. For training images, we apply two off-the-shelf pose estimators~\cite{li2020self} and human parsers~\cite{gong2017look} to extract the human body keypoints and semantic maps, respectively. The detailed process is as follows:

\textbf{Human pose embedding.} Following several off-the-shelf virtual try-on methods~\cite{han2018viton, wang2018toward,yu2019vtnfp,dong2019towards,zheng2019virtually,han2019clothflow}, we use the pose estimator~\cite{cao2017realtime} to extract the pose of each person image. Then, we obtain the coordinates of 18 human body keypoints from each person image and convert them to an 18-channel heatmap. Each channel of the heatmap corresponds to a human pose keypoint. We use each keypoint as the center of the circle to draw a circle with a radius of 4 pixels. The values in each circle are all ones, and the parts outside the circle are all zeros. In this way, we obtain the representation of the human pose embedding.  

\textbf{Human semantic map.} 
Inspired by two semantically guided virtual try-on approaches~\cite{yu2019vtnfp, dong2019towards}, we extract human semantic maps of training images by using the existing human parser~\cite{gong2017look}. Each semantic map contains 20 labels that correspond to different parts of the human body.
For intractable human body parts, such as the head (including the face and hair regions), it preserves characteristic personal identity information. The extracted information provided an additional supervision signal for training our network.

\subsection{Semantic Prediction Module}
\label{3.2}
To precisely locate the clothes region of the generated person image and alleviate the mismatching between the target clothes and the generated human body, we introduce a semantic prediction module (SPM). As shown in Fig. \ref{fig:overview} {a}, SPM consists of two sequential processes and can be optimized in one step. The first process is target semantic map prediction, and the other is a target clothing mask and coarse try-on result generation. First, given a source human image $I_{s}$ and its corresponding semantic map $S_{s}$, a target in-shop clothes $C$, and a target human pose $P_{t}$, the first process of SPM aims to predict the target human semantic map $\hat{S_{t}}$ conditioned on the source semantic map $ S_{s}$, the target clothes $C$, and the target pose $P_{t}$.
Second, we combine the predicted semantic map $\hat{S_{t}}$ with the desired clothes $C $, the target pose $ P_{t}$, and the source image without clothes $I^{w/o. C}_{s}$ and then feed it into the second process of SPM to generate the coarse try-on image and the predicted clothes mask. Specifically, each process of the SPM is based on the conditional generative adversarial network (cGAN)~\cite{mirza2014conditional}. We adopt a ResNet-like network to replace the U-Net~\cite{ronneberger2015u} structure as the generator $G$, and the multiscale discriminator (PatchGAN)~\cite{wang2018high} is applied as the discriminator $D$. SPM contains two generators (\ie, $G_{1}$ and $G_{2}$) and two discriminators (\ie, $D_{1}$ and $D_{2}$). $G_{1}$ produces the predicted semantic map of the target person that makes discriminator $D_{1}$ indistinguishable from the real image. Similarly, the role of generator $G_{2}$ is to generate the realistic coarse result and the predicted clothing mask. At the same time, discriminator $D_{2}$ attempts to distinguish real images from the results generated by $G_{2}$. The network structure of generators and discriminators can be found in Table \ref{tab:architecture}.

\textbf{Target semantic map prediction.} 
We first define the head part as $A_{s}=\{M^{f}_{s}, M^{h}_{s}\}$, which is composed of the face mask $M^{f}_{s}$ and the hair mask $M^{h}_{s}$. $A_{s}$ means that all the masks in $A$ are obtained from the source semantic map $S_{s}$. Next, we remove the hair, face, and clothing areas from the source semantic map $S_{s}$ to obtain the reference semantic map $ S^\prime_{s}$. Finally, we binarize the desired clothing image $C$ to obtain the mask of desired clothing $M^{C}$.
As shown in Fig.~\ref{fig:overview} (a), we combine the head part $A_{s}$ with the reference semantic map $ S^\prime_{s}$, the mask of desired clothes $M^{C}$, and the target pose $P_{t}$ as the input of the target semantic map prediction. Therefore, the predicted semantic map can be formulated as $\hat{S_{t}} = G_{1}( S^\prime_{s},A_{s},M^{C},P_{t})$. To encourage the generated semantic map to be indistinguishable from the ground-truth semantic map, we introduce the spatial matching adversarial loss~\cite{goodfellow2014generative} as follows:
\begin{align}
\begin{split}
   L^{s}_{adv}=~&\mathbb{E}[\log D_{1}(S_{t},S^\prime_{s},A_{s},M^{C},P_{t})]\\ 
   +~&\mathbb{E}[\log(1\!-\!D_{1}(G_{1}(S^\prime_{s},A_{s},M^{C}\!,\!P_{t}),S^\prime_{s},A_{s},M^{C},P_{t})].
\end{split}
\end{align}

In addition, to generate high-quality target semantic maps $\hat{S_{t}}$, we utilize the focal loss~\cite{gong2017look} on pixelwise segmentation. 
In addition, following \cite{yan2017skeleton,dong2019towards}, we also adopt the pixelwise $\mathcal{L}_{1}$-norm between the predicted semantic map and the target semantic map to push the generator $ G_{1}$ to produce smoother results.
Therefore, the objective function to generate the target semantic map can be formulated as:
\begin{align}
L_{seg} = L^{s}_{adv} + \lambda_{1}L^{s}_{fl} + L^{s}_{recon},
\end{align}
where $L^{s}_{fl}$ denotes the focal loss between $\hat{S_{t}}$ and $S_{t}$, $L^{s}_{recon}$ denotes the pixelwise $\mathcal{L}_{1}$-norm between the predicted semantic map $ \hat{S_{t}}$ and the target semantic map $ S_{t}$, and the hyperparameter $\lambda_{1}$ controls the weights of the focal loss.

\textbf{Target clothing mask prediction and coarse result generation.}
As shown in Fig.~\ref{fig:overview} (a), after obtaining the predicted semantic map $\hat{S_{t}}$ from the first process of SPM, we first remove the clothe region of $\hat{S_{t}}$ to obtain the incomplete predicted semantic map $ \hat{S^{'}_{t}}$ and then combine $ \hat{S^{'}_{t}}$ with the target pose $P_{t}$, the in-shop clothes $C$, and the source image without clothes $I^{w/o. C}_{s}$ as the input of the second process of SPM (\ie, the input of the generator $ G_{2}$). Then, the generator $ G_{2}$ produces both the coarse try-on image $ I_{p}$ and the predicted clothing mask $ M^{c}_{p}$.
To allow the generator to produce the photo-realistic coarse try-on image $I_{p}$, we also define the adversarial loss $L^{p}_{adv}$ as follows:
\begin{align}
\begin{split}
   L^{p}_{adv}=~&\mathbb{E}[\log D_{2}(I_{t},\hat{S_{t}},C,P_{t},I^{w/o. C}_{s})]  \\
   +~&\mathbb{E}[\log(1-D_{2}(I_{p},\hat{S_{t}},C,P_{t},I^{w/o. C}_{s})].
\end{split}
\end{align}

 Following several start-of-the-art virtual try-on methods~\cite{han2018viton,wang2018toward,dong2019towards,yu2019vtnfp,han2019clothflow,zheng2019virtually,HanYang2020}, we also adopt the perceptual loss~\cite{johnson2016perceptual} to enforce the generator $G_{2}$ to synthesize photo-realistic try-on images. The perceptual loss between $I_{p}$ and $I_{t}$ can be defined as:
\begin{align}
\label{eq5}
\begin{split}
    L^{p}_{perc} = \mathbb{E}[\sum_{i=0}^{5} \sigma_{i} \| \phi_{i}(I_{p}) - \phi_{i}( I_{t}) \|_{1}],
\end{split}
\end{align}
where $\phi_{i}(I_{.})$ denotes the $i$-th layer feature map of the image $I_{.}$ in the visual perception network $\phi$. We apply the pretrained VGG-19~\cite{Simonyan15} network as $\phi$. Here, five activations are utilized to calculate the perceptual loss. The hyperparameters $\sigma_{i}$ control the weights of the $i$-th layer to the term in Eq.~\ref{eq5}. We apply the pixelwise $\ell_{1}$ loss to guide the target clothing mask prediction and the coarse result generation. Therefore, to minimize the distance between the generated image and the ground-truth image at the pixel level, we formulate the loss function as:
\begin{align}
\label{eq6}
\begin{split}
L^{p}_{recon} = \mathbb{E}[\| I_{p} - I_{t}\|_{1}] +
\mathbb{E}[\| M^{c}_{p} -  M^{c}_{t}\|_{1}],
\end{split}
\end{align}
where $M^{c}_{p}$ represents the predicted clothing mask, and $M^{c}_{t}$ denotes the clothing region mask extracted from the target human image $I_{t}$. The objective function to generate the coarse try-on result and target clothing mask can be formulated as:
\begin{align}
\label{eq7}
    L_{prd =  L^{p}_{adv} +\lambda_{2}(L^{p}_{recon} + L^{p}_{perc}),}
\end{align}
where $\lambda_{2}$ control weights of $L^{p}_{recon}$ and $ L^{p}_{perc} $. Then, the full objective function of the SPM can be defined as:
\begin{align}
\begin{split}
\label{eq72}
    L_{spm} = ~& L_{seg} + L_{prd} \\
    =~& L^{s}_{adv} + \lambda_{1}L^{s}_{fl} + L^{s}_{recon} \\
    +~& L^{p}_{adv} +\lambda_{2}(L^{p}_{recon} + L^{p}_{perc}),
\end{split}
\end{align}

\subsection{Clothes Warping Module}
\label{3.3}
The clothes warping module (CWM) aims to fit the desired clothes onto the target person according to the given pose while preserving the texture the of clothes. Most existing works~\cite{wang2018toward, yu2019vtnfp, dong2019towards,zheng2019virtually} directly utilize a geometric matching module (GMM)~\cite{rocco2017convolutional} to estimate the parameters of the thin-plate spline~(TPS) used to warp clothes. This strategy is applicable when the texture of the clothes is monotonous and the target pose is fixed. However, when dealing with complex cases (\emph{e.g.,} the desired clothes with complex texture and the target person under diversity poses), it might lead to misalignment between clothes and the human body and blurred results. To address the challenges mentioned above, we introduce the conductible cycle consistency loss, which effectively aligns the desired clothing with a given pose. 

As shown in Fig.~\ref{fig:overview} (b), we first apply the geometric matching network~\cite{rocco2017convolutional} to estimate the TPS transformation parameters $\theta_{1}$ between the mask of desired clothes $M^{C}$ and the predicted clothes mask $M^{c}_{p}$. The mask of desired clothes $M^{C}$ is then warped using the transformation parameters $\theta_{1}$ to align it with the predicted clothes mask $M^{c}_{p}$. We denote the operation of TPS transformation as $\mathcal{T}_{\theta}$, and then the warped mask of desired clothes $M^{C}_{w}$ can be represented as $M^{C}_{w} = \mathcal{T}_{\theta_{1}}(M^{C})$. Next, the second geometric matching network is adopted to estimate the TPS transformation parameters $\theta_{2}$ between the warped mask of desired clothes $M^{C}_{w}$ and the mask of desired clothes $M^{C}$. We use the TPS transformation parameters $\theta_{2}$ to warp $M^{C}_{w}$ back to the original mask of desired clothes $M^{C}$. We denote the output of the second geometric matching network as $\hat{M^{C}} = \mathcal{T}_{\theta_{2}}(M^{C}_{w})$. 

\textbf{Conductible cycle consistency loss.}
A straightforward solution is to train the geometric matching network by directly applying the pixel-level $\mathcal{L}_{1}$-norm to encourage $M^{C}_{w}$ to approximate $M^{c}_{p}$ and $\hat{M^{C}}$ to approximate $M^{C}$ to obtain the TPS transform parameters for the desired clothing deformation.
However, using the self-reconstruction approach to estimate TPS parameters between the given clothing mask and the target clothing mask can only roughly align the shape of the given clothing mask with the target clothing mask. Applying the estimated parameters to warp the desired clothes directly cannot preserve the details of the clothes well. Based on these observations, we introduce the conductible cycle consistency loss for two goals. One is to match the shape of the deformed desired clothes with the target person image. The other is to maintain the characteristics of the desired clothes in the clothing region of the generated try-on image.
Specifically, we adopt the estimated parameters $\theta_{1}$ to warp the desired clothes $C$ to obtain the warped clothes $C_{w} = \mathcal{T}_{\theta_{1}}(C)$ and then use the estimated parameter $\theta_{2}$ to warp $C_{w}$ to produce $\hat{C} = \mathcal{T}_{\theta_{2}}(C_{w})$. Thus, we formulate the conductible cycle consistency loss as:
\begin{align}
\label{eq8}
    L_{cond} =  \mathbb{E}[\| C_{w} - C_{t}\|_{1}] +
    \mathbb{E}[\| \hat{C} - C\|_{1}],
\end{align}
where $C_{t}$ denotes the ground-truth clothing region extracted from the target image $I_{t}$. Then, the full objective function of the CWM can be defined as:
\begin{align}
 L_{cwm} = \lambda_{3}L_{cond},
\end{align}
where $\lambda_{3}$ controls weights of the conductible cycle consistency loss.

\textbf{Discussion.}  
The existing methods~\cite{han2018viton, wang2018toward,yu2019vtnfp,dong2019towards,zheng2019virtually} adopt a separate clothing deformation module to calculate the TPS transformation parameters between the desired clothing and the target person image. The TPS transformation parameters are given by the pretrained clothing deformation module, which cannot be dynamically adjusted for these parameters and leads to error stacking. In contrast to previous works~\cite{han2018viton, wang2018toward,yu2019vtnfp,dong2019towards,zheng2019virtually}, we adopt an indirect approach to train the geometric matching network. Concretely, we first fed the mask of desired clothes and the predicted clothing mask of the target person image into the geometric matching network and then used the estimated TPS transformation parameters to warp the desired clothes. The pixelwise $\mathcal{L}_{1}$ loss between the warped desired clothes and the ground-truth clothes extracted from the target person image is the constraint of the geometric matching network. Furthermore, we adopt the pixelwise $\mathcal{L}_{1}$-norm between $\hat{C}$ and the desired clothes $C$ to encourage the warped mask of desired clothes $M^{C}_{w}$ to return to the original mask of desired clothes $M^{C}$. This indirect constraint strategy can not only accurately achieve geometric deformation between the desired clothes and the target person image under an arbitrary pose but also preserve rich details of the clothing in the generated person image. The ablation study verifies the effectiveness of the conductible cycle consistency loss.

\subsection{Try-on Synthesis Module}
\label{3.4}
After producing the coarse try-on result $I_{p}$ by SPM, we first randomly erase a part of $I_{p} $ to obtain the incomplete coarse result $I^{'}_{p} $, and then we combine $I^{'}_{p} $ and the deformed desired clothes $ C_{w}$ obtained by CWM and the target pose $ P_{t}$. Next, we directly fed them into the TSM to generate the final virtual try-on image $\hat{I_{t}}$. As shown in Fig.~\ref{fig:overview} (c), we use the generator $G_{3}$ to produce a rendered result $I_{r}$ and a composition mask $\mathcal{M}$ at the same time. The final virtual try-on result can be formulated as follows:
\begin{align}
\label{eq9}
    \hat{I_{t}} = C_{w} \odot \mathcal{M} + I_{r}\odot (1-\mathcal{M}),
\end{align}
where $(I_{r}, \mathcal{M}) = G_{3}(I^{'}_{p}, C_{w}, P_{t})$. $\odot$ represents the elementwise matrix multiplication, and the clothes part in the final try-on image can be denoted as $\hat{C_{t}} = C_{w} \odot \mathcal{M}$. The value of each element in $\mathcal{M}$ is between 0 and 1. Following several start-of-the-art virtual try-on methods~\cite{wang2018toward,yu2019vtnfp,zheng2019virtually,HanYang2020}, we apply both the self-reconstruction loss and the perceptual loss~\cite{johnson2016perceptual} to enforce the generated image $\hat{I_{t}}$ to approximate the target image $I_{t}$. We define the full reconstruction loss as:
\begin{align}
\begin{split}
\label{eq10}
    L^{t}_{recon} = \alpha_{1} \mathbb{E}[\| \hat{I_{t}} - I_{t}\|_{1}]) + \alpha_{2} \mathbb{E}[\| 1 - \mathcal{M}\|_{1}],
\end{split}
\end{align}
 where we adopt the second term in Eq.~\ref{eq10} as the regularization to constrain the generation of composition mask $\mathcal{M}$. We set $\alpha_{1} = 2$ and $\alpha_{2} = 0.5$. Additionally, the perceptual loss between $\hat{I_{t}}$ and $I_{t}$ can be formulated as follows:
\begin{align}
\label{eq10-2}
     L^{t}_{perc} = \mathbb{E}[\sum_{i=0}^{5} \sigma_{i} \| \phi_{i}(\hat{I_{t}}) - \phi_{i}( I_{t}) \|_{1}].
\end{align}

\begin{figure}[t]
  \centering
  \includegraphics[width=\linewidth]{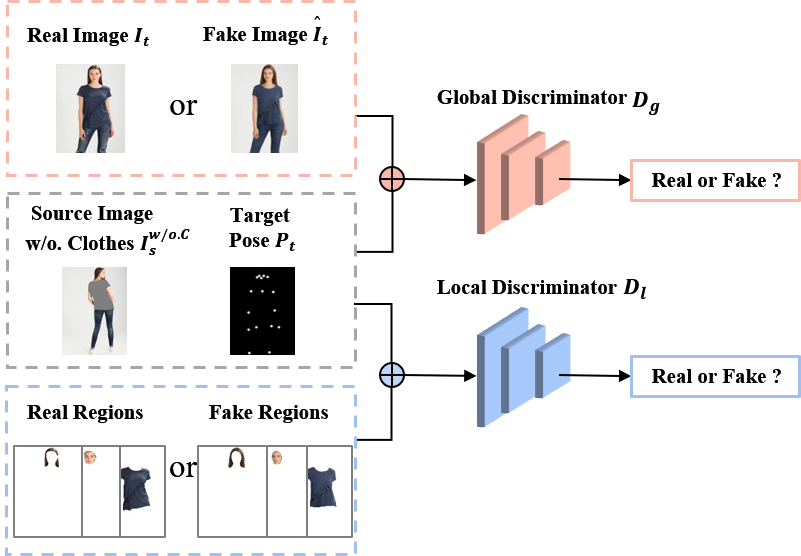}
  \vspace{-.5em}
  \caption{Diagram of global and local discriminators. More details can be found in Table~\ref{tab:architecture}.}
  \label{fig:discriminator}
\end{figure}

\textbf{Global and local adversarial loss.} In the real scenario, it is difficult to obtain images of the same person wearing the desired clothes in arbitrary poses. Therefore, we cannot retain parts outside the clothing area following many existing virtual try-on methods based on a fixed pose.
In this case, to generate the photo-realistic try-on image,
we employ a global adversarial loss and a local adversarial loss in TSM. The diagram of global and local discriminators is shown in Fig.~\ref{fig:discriminator}.
Specifically, we apply the global adversarial loss $L^{g}_{adv}$ to enforce the generator $G_{3}$ to synthesize sharp virtual try-on images with global consistency. Furthermore, the local adversarial loss $L^{l}_{adv}$ is adopted to refine the face area of the final result with local consistency.
The global and local adversarial loss can be formulated as follows:
\begin{align}
\label{eq11}
\begin{split}
   L^{g}_{adv}=~&\mathbb{E}[\log D_{g}(I_{t},C_{t},P_{t},I^{w/o.C}_{s})]\\ +~&\mathbb{E}[\log(1-D_{g}( \hat{I_{t}}, \hat{C_{t}},P_{t},I^{w/o.C}_{s})],
\end{split}
\end{align}
\begin{align}
\label{eq11-2}
\begin{split}
   L^{l}_{adv}=~&\mathbb{E}[\log D_{l}(I^{f}_{t},I^{h}_{t},C_{t},P_{t},I^{w/o.C}_{s})]\\
   +~&\mathbb{E}[\log(1-D_{l}(\hat{I^{f}_{t}},\hat{I^{h}_{t}},\hat{C_{t}},P_{t},I^{w/o.C}_{s})],
\end{split}
\end{align}
where we extract the face region $I^{f}_{t}$ and the hair region $I^{f}_{t}$ from the ground-truth image $I_{t}$. In the same way, we also extract the face region $\hat{I^{f}_{t}}$ and the hair region $\hat{I^{h}_{t}}$ from $\hat{I_{t}}$. $D_{g}$ denotes the global discriminator, and $D_{l}$ denotes the local discriminator. $D_{g}$ and $D_{l}$ share the same structure.

\textbf{Face identity loss.} After adopting the global and local adversarial loss, our method can produce high-quality fitting images while making the face region look natural. Furthermore, to enforce that the face region cropped from the generated images remains similar to the face region of the related ground-truth images, we introduce the face identity loss.  
\begin{align}
\label{eq12}
    L_{id} = \mathbb{E}[\| \mathcal{F}(\hat{I^{F}_{t}})- \mathcal{F} (I^{F}_{t})\|_{1}] ,
\end{align}
where $ \mathcal{F}$ denotes the pretrained SphereFaceNet~\cite{liu2017sphereface}. $\hat{I^{F}_{t}}$ represents the face region extraction guided by the predicted semantic map, including the face part and the hair part.
Then, we formulate the full objective function of the TSM as:
\begin{align}
\label{eq13}
   L_{tsm} = \lambda_{4} (L^{t}_{recon} + L^{t}_{perc})+ L^{g}_{adv} + L^{l}_{adv} + \lambda_{5} L_{id},
\end{align}
where the hyperparameter $\lambda_{4}$ controls weights of the reconstruction loss $L^{t}_{recon}$ and the perceptual loss $L_{perc}(\hat{I_{t}}, I_{t}))$, and $\lambda_{5}$ controls weights of the face identity loss $ L_{id}$.

\subsection{Optimization}

\textbf{Objective function.} Taking all of the above loss functions into consideration, we formulate the total objective function as:

\begin{align} 
\begin{split}
L_{total}&=
\begin{matrix} \underbrace{L^{s}_{adv} \!+\! \lambda_{1}L^{s}_{fl}\!+\!L^{s}_{recon}\!+\! L^{p}_{adv}\!+\!\lambda_{2}(L^{p}_{recon} \!+\! L^{p}_{perc})} \\ \footnotesize{\text{SPM}}\end{matrix} \\
&~\!+\!\begin{matrix} \underbrace{\lambda_{3}L_{cond}} \\ \footnotesize{\text{CWM}}\end{matrix}
\!+\!\begin{matrix} \underbrace{  \lambda_{4} (L^{t}_{recon}\!+\!L^{t}_{perc})\!+\! L^{g}_{adv}\!+\!L^{l}_{adv} \!+\!\lambda_{5} L_{id}} \\ \footnotesize{\text{TSM}}\end{matrix}.
\end{split}
\label{loss1}
\end{align}

Our ultimate goal is to solve:
\begin{align} 
\begin{split}
G^{*}, \mathcal{T}^{*} =\arg\underset{G_{1},G_{2}, \mathcal{T}_{\theta_{1}},\mathcal{T}_{\theta_{2}},G_{3}}\min\ \underset{D_{1},D_{2},D_{g},D_{l}}\max\ L_{total}.
\end{split}
\label{loss2}
\end{align}

\textbf{End-to-end training.} In this article, we divide the training process of the proposed method into three steps:
(1) We first separately train the semantic prediction network to obtain the preliminary semantic prediction map, and this process corresponds to the first process of the semantic prediction module.
(2) Then, we use the semantic prediction map generated from the fixed pretrained semantic prediction network as guidance for the subsequent steps, including rough result generation, clothing area prediction, target clothing deformation, and refined try-on generation.
(3) Finally, we jointly train three submodules of the proposed method to synthesize the final virtual try-on image. This step can alleviate the impact of inaccurate semantic maps and improve the quality of the generated results.

\textbf{Differences from MG-VTON~\cite{dong2019towards}.} 
There are significant differences between MG-VTON and our approach.

(1) MG-VTON adopts a multistage framework, and each stage independently implements a distinct task. In particular, MG-VTON applies different modules to achieve the desired clothing deformation, coarse result generation, and final result refinement separately. Instead, we employ end-to-end training to encourage the generator to produce realistic virtual try-on results. Specifically, we dynamically update the parameters of each process, including the desired clothing deformation, the coarse result generation, and the final result refinement. 
(2) Both MG-VTON and SPG-VTON use a semantic prediction module to predict the target semantic map (\ie, the first process of SPM in SPG-VTON and the conditional human parsing network in MG-VTON). However, unlike MG-VTON, SPM in SPG-VTON consists of one extra clothes mask prediction. Therefore, SPG-VTON can accurately locate the target clothing area by combining the predicted target semantic map and the predicted target clothing mask.
(3) MG-VTON divides the coarse result generation and the composition mask production into two independent steps. In contrast to this technique, we concatenate the coarse result, the warped clothes, and the target pose as the input of $G_{3}$ to produce a rendered result and a composition mask at the same time. Then, we combine the warped clothes, the rendered result, and the composition mask to synthesize the final result. These differences make our method generate images with improved qualities in both qualitative and quantitative evaluations, which is demonstrated in Fig.~\ref{fig:compare}, TABLE \ref{tab:ssim_is} and TABLE \ref{tab:vsmg}.

\section{Experiments}
\subsection{Dataset}
We perform experiments on the MPV dataset~\cite{dong2019towards}, which is the largest multi-pose virtual fitting dataset available. The MPV dataset consists of 35,687 person images and 13,524 clothing images collected from the internet, with a resolution of $256\times192$. For each in-shop item of clothing, the dataset contains multiple images of the same person wearing the given in-shop clothing in different poses. The MPV dataset contains 62,780 three-tuples, including 52,236 training sets and 10,544 test sets. Each input data point is a three-tuple composed of two human images and one clothes image. The two person images wear the clothes in the clothes image but with different poses. Moreover, we also conduct experiments on the DeepFashion dataset (\emph{In-shop Clothes Retrieval Benchmark})~\cite{liu2016deepfashion} to verify the effectiveness of the proposed method. Following the setting in MG-VTON~\cite{dong2019towards}, we collect 10,000 pairs (the same person in different poses) from DeepFashion and randomly select in-shop clothes from the test set of the MPV dataset.

\begin{figure}[t]
  \centering
  \includegraphics[width=\linewidth]{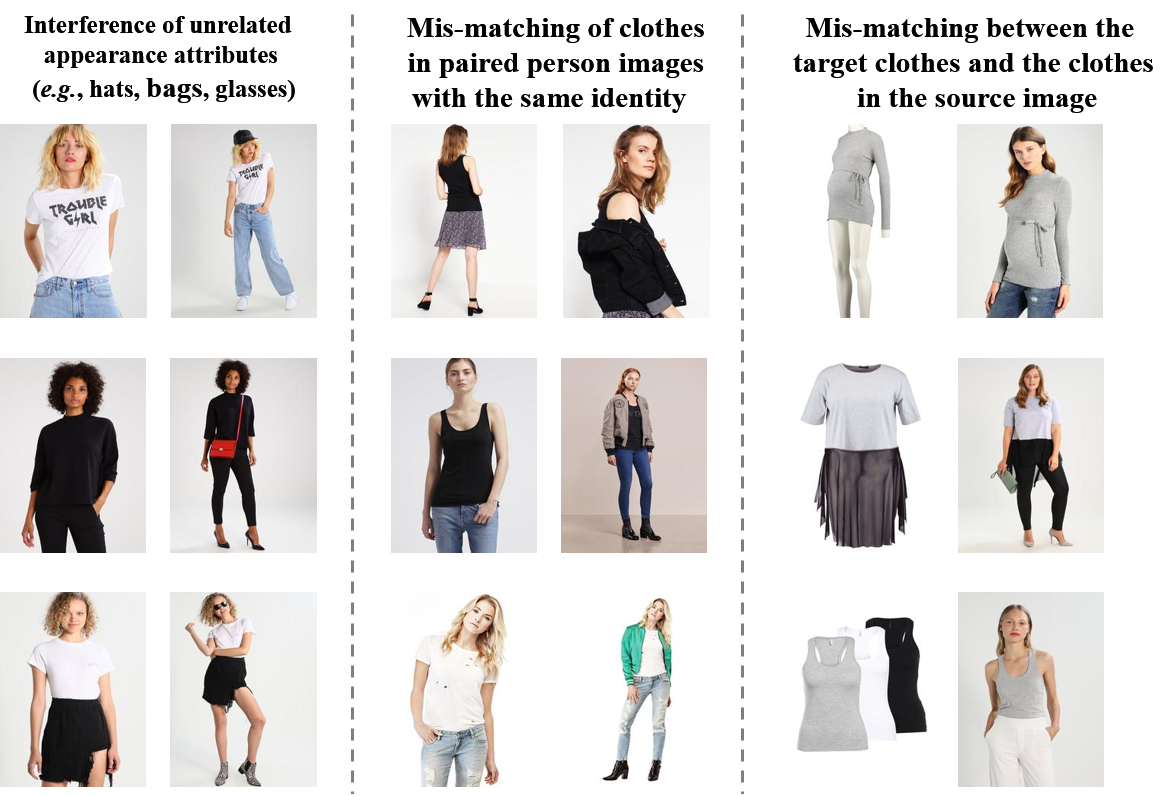}
  \vspace{-1em}
  \caption{Examples of noisy images in the training set and the test set. We observe the three typical kinds of noise existing in the training set and the test set, including interference of unrelated appearance attributes (\eg, hats, bags, glasses), mismatching of clothes in paired person images with the same identity, and mismatching between the target clothes and the clothes in the source image. These noise demands strong robust ability of the proposed algorithm during both training and testing.
}
  \label{fig:noise}
\end{figure}

\textbf{Noise in the training set and the test set.} For the virtual try-on task, the input three-tuple consists of two paired person images and one in-shop clothing image. When one of the following conditions occurs, the input three-tuple can be regarded as "noise data."
(1) One person image contains attributes that do not exist in the other person image, and these attributes are limited to glasses, bags, hats, scarves, necklaces, gloves, coats, upper clothes, and background. (2) The given clothes are different from the clothes in the paired person images. Since MPV does not provide labels for the noise data, in this case, to accurately obtain the noise data in the training set, we manually filter it and obtain 8,740 sets of noise data, accounting for approximately 16.73\% of the training set.
We show three typical kinds of noise in both the training set and the test set in Fig.~\ref{fig:noise}, including interference of unrelated appearance attributes (\eg, hats, bags, glasses), mismatching of clothes in paired person images with the same identity, and mismatching between the target clothes and the clothes in the source image. It is challenging to train a robust virtual fitting system based on these noisy images.

\subsection{Evaluation Metrics}

\textbf{Structural SIMilarity (SSIM).} SSIM~\cite{wang2004image} is widely used to evaluate the similarity of generated images in GAN-based methods. In this work, we adopt the SSIM metric to measure the similarity between the generated image and the real image. Higher scores indicate that the generated image is closer to the ground-truth image.

\textbf{Inception Score (IS).} IS~\cite{salimans2016improved} is a general metric used to estimate the quality of the synthesis image. In this work, we apply IS to evaluate the quality of generated images by our method. Notably, all generated images used to calculate IS have no corresponding ground-truth images.
To evaluate the quality of specific regions (\ie, clothing regions, face regions) in the final virtual try-on image, we also calculate Mask-SSIM and Mask-IS for the extracted regions.

\subsection{Implementation Details}
\textbf{Architecture.} Here, we provide details about the network architecture of three submodules in SPG-VTON. Specifically, generators $ G_{1}$, $G_{2}$, and $G_{3}$ adopt the same structure, which is a ResNet-like architecture. The generated result of $G_{2}$ and $G_{3}$ is a 4-channel tensor that could be split into a 1-channel mask and a 3-channel RGB image. We show the detailed network structure of generators and discriminators in TABLE~\ref{tab:architecture}. In addition, all the discriminators in the proposed method apply the multiscale structure from pix2pixHD~\cite{wang2018high}. Additionally, we use instance normalization~\cite{ulyanov2016instance} in both generators and discriminators. We employ ReLU and LeakyReLU activation functions in the generator and discriminator, respectively. Following existing works~\cite{zheng2019joint,hu2020unsupervised}, we adopt LSGAN~\cite{mao2017least} for all adversarial losses in our method, and a gradient punishment strategy \cite{mescheder2018training} is also used to stabilize the training process. 

\setlength{\tabcolsep}{0.6mm}{
\begin{table}[t] 
\setlength{\abovecaptionskip}{0pt}%
\setlength{\belowcaptionskip}{2pt}%
\begin{center}
\caption{The network architecture of our generators and discriminators, where K, S, P, and A denote the kernel size, stride size, padding size, and the activation function, respectively. \textbf{IN} represents the input channels, and \textbf{OUT} denotes the output channels. The generators $G_{1}$, $G_{2}$, $ G_{3}$ adopt the same structure while the input channels and output channels are different. Besides, discriminators $D_{1}$, $D_{2}$, $ D_{g}$, $ D_{l}$ share the same structure while the input channels are different.
}
\vspace{-.2em}\scriptsize
\begin{tabular}{|l|c|c|c|}
    \hline
    \multicolumn{4}{|c|}{Generator $ G_{.} $}\\
    \hline
    \hline
    Layer & Input & Output & K \& S \& P \& A\\
    \hline
     Conv1 & \textbf{IN} $\times$256$\times$192 & 64$\times$256$\times$192 & 3$\times$3, 1, 1, ReLU\\
    \hline
     Conv2 & 64$\times$256$\times$192 & 128$\times$256$\times$192 & 3$\times$3, 1, 1, ReLU\\
    \hline
     Conv3 & 128$\times$256$\times$192 & 128$\times$128$\times$96 & 3$\times$3, 2, 1, ReLU\\
    \hline
     Resblock & 128$\times$128$\times$96 & 128$\times$128$\times$96 & 3$\times$3, 3$\times$3, 1, 1, ReLU\\
    \hline
     Conv4 & 128$\times$128$\times$96 & 256$\times$64$\times$48 & 3$\times$3, 2, 1, ReLU\\
    \hline
     Resblock & 256$\times$64$\times$48 & 256$\times$64$\times$48 & 3$\times$3, 3$\times$3, 1, 1, ReLU\\
    \hline
     Conv5 & 256$\times$64$\times$48 & 512$\times$32$\times$24 & 3$\times$3, 2, 1, ReLU\\
    \hline
     Resblock $\times$ 4 & 512$\times$32$\times$24 & 512$\times$32$\times$24 & 3$\times$3, 3$\times$3, 1, 1, ReLU\\
    \hline
     Upsample & 512$\times$32$\times$24 & 512$\times$64$\times$48 & - \\
    \hline
     Conv6 & 512$\times$64$\times$48 & 256$\times$64$\times$48 & 3$\times$3, 1, 1, ReLU\\
    \hline
     Upsample & 256$\times$64$\times$48 & 256$\times$128$\times$92 & - \\
    \hline 
     Conv7 & 256$\times$128$\times$92 & 128$\times$128$\times$92 & 3$\times$3, 1, 1, ReLU\\
    \hline
     Upsample & 128$\times$128$\times$92 & 128$\times$256$\times$192 & - \\
    \hline
     Conv8 & 128$\times$256$\times$192 & 64$\times$256$\times$192 & 3$\times$3, 1, 1, ReLU\\
    \hline
     Conv9 & 64$\times$256$\times$192 & \textbf{OUT} $\times$256$\times$192 & 3$\times$3, 1, 1, ReLU\\
    \hline 
     Conv10 & \textbf{OUT} $\times$256$\times$192 & \textbf{OUT} $\times$256$\times$192 & 3$\times$3, 1, 1, Tanh\\
    \hline
     Conv11 & \textbf{OUT} $\times$256$\times$192 & \textbf{OUT} $\times$256$\times$192 & 1$\times$1, 1, 0, None\\

    \hline
    \hline
    \multicolumn{4}{|c|}{Discriminator $ D_{.} $}\\
    \hline
    \hline
    Layer & Input & Output & K \& S \& P \& A\\
    \hline
     Conv1 & \textbf{IN} $\times$256$\times$192 & 32$\times$256$\times$192 & 3$\times$1, 1, 0, LeakyReLU\\
    \hline
     Conv2 & 32$\times$256$\times$192 & 32$\times$256$\times$192 & 3$\times$3, 1, 1, LeakyReLU\\
    \hline
     Conv3 & 32$\times$256$\times$192 & 32$\times$128$\times$96 & 3$\times$3, 2, 1, LeakyReLU\\
    \hline
     Conv4 & 32$\times$128$\times$96 & 32$\times$128$\times$96 & 3$\times$3, 2, 1, LeakyReLU\\
    \hline
     Conv5 & 32$\times$128$\times$96 & 64$\times$64$\times$48 & 3$\times$3, 2, 1, LeakyReLU\\
    \hline
     Conv6 & 64$\times$64$\times$48  & 1 $\times$64$\times$48 & 1$\times$1, 1, 0, None\\
    \hline

\end{tabular}
  \label{tab:architecture}
\end{center}
\end{table}
}

\begin{table}[t]
\centering
\caption{\textbf{Quantitative results.} Comparison results in terms of SSIM and IS on both MPV and DeepFashion. $\uparrow$ denotes that higher scores are better.}
\vspace{-.5em}\scriptsize
\resizebox{0.9\columnwidth}{!}{
\setlength{\tabcolsep}{4pt}
\begin{tabular}{l|cc|cc}
	\toprule
       \multirow{2}{*}{Method}  & \multicolumn{2}{c|}{MPV} &  \multicolumn{1}{c}{DeepFashion} \\
     \cmidrule{2-4} 
         & SSIM $\uparrow$ & IS $\uparrow$ & IS $\uparrow$ \\
    \midrule
            Real data & 1.000 & 3.391 $\pm$ 0.024 & 3.332 $\pm$ 0.126 \\
        \midrule   
        VITON~\cite{han2018viton} & 0.639 & 2.394 $\pm$ 0.205 & 2.302 $\pm$ 0.116 \\
        CP-VTON~\cite{wang2018toward} & 0.705 & 2.519 $\pm$ 0.107 & 2.459 $\pm$ 0.212	\\ 
        MG-VTON~\cite{dong2019towards} & 0.744 & 3.154 $\pm$ 0.142 &  3.030 $\pm$ 0.057	\\
        SPG-VTON (Ours)	& \textbf{0.752} & \textbf{3.243 $\pm$ 0.127} &   \textbf{3.124 $\pm$ 0.027}	\\
    \bottomrule
\end{tabular}
}
\label{tab:ssim_is}
\end{table}

\begin{table}[t]
\centering
\caption{\textbf{SPG-VTON \emph{vs.} MG-VTON~\cite{dong2019towards}.} Quantitative comparison results in terms of Mask-SSIM and Mask-IS on the part of MPV. $\uparrow$ denotes that higher scores are better. Notably, the source image and target pose in each tuple in the test set have the same identity, so we calculate Mask-SSIM values of the facial area and the area unrelated to clothes between the generated image and the original image that provides the target pose.
}
\vspace{-.5em}
\resizebox{0.9\columnwidth}{!}{
\setlength{\tabcolsep}{3pt}
\begin{tabular}{l|c|c|c|c}
    \toprule
	   \multirow{2}{*}{Method} &  \multicolumn{2}{c|}{ Mask-SSIM $\uparrow$}& \multicolumn{2}{c}{Mask-IS $\uparrow$}  \\
      \cmidrule{2-5} 
         & Face  &  w/o Clothes & Face  &  Clothes\\
    \midrule 
        Real data & 1.000 & 1.000 & 
        1.848 $\pm$ 0.137 & 
        3.711 $\pm$ 0.237 \\
    \midrule
        MG-VTON~\cite{dong2019towards} & 
        0.717 $\pm$ 0.094 & 
        0.722 $\pm$ 0.045 & 
        1.419 $\pm$ 0.061 & 
        3.270 $\pm$ 0.409 \\
        SPG-VTON (Ours) & 
        \textbf{0.737 $\pm$ 0.091} & \textbf{0.745 $\pm$ 0.042} & \textbf{1.584 $\pm$ 0.053} & \textbf{3.660 $\pm$ 0.327}\\
    \bottomrule
\end{tabular}
}
\label{tab:vsmg}
\end{table}

\begin{figure*}[t]
  \centering
  \includegraphics[width=\linewidth]{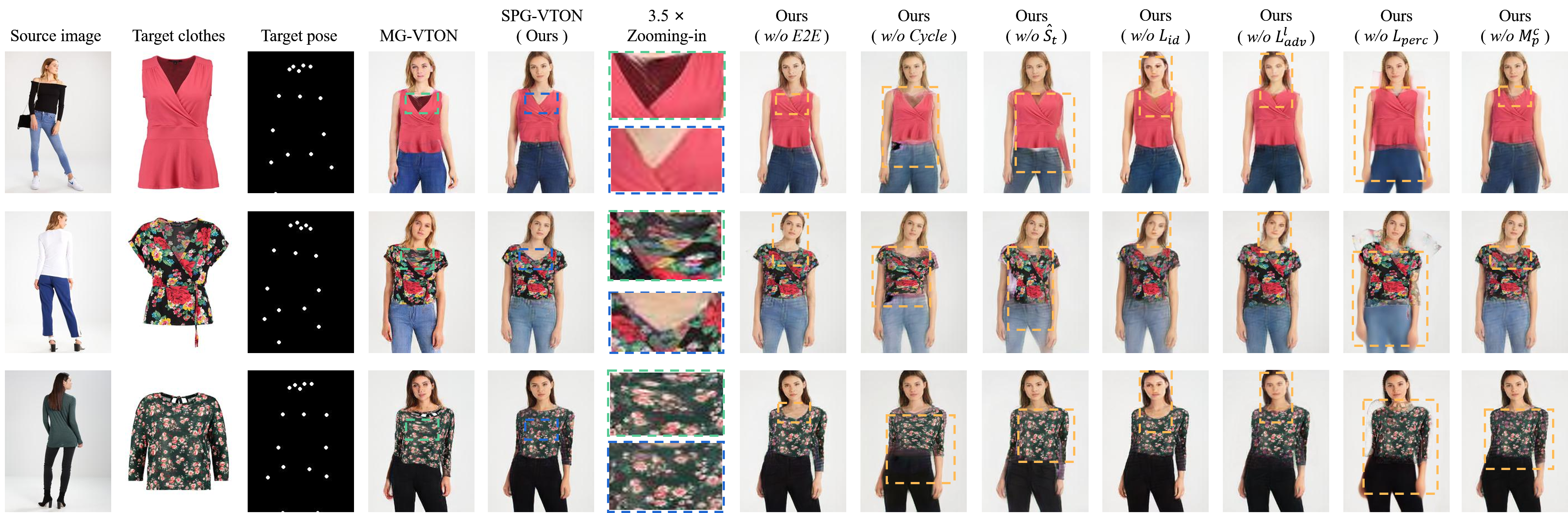}
  \vspace{-.5em}
  \caption{Visualized comparison between MG-VTON~\cite{dong2019towards} and several variants of our method on the MPV dataset. To make the comparison clearer, we use the green dashed box and the yellow dashed box to cut out the local areas of generated images by MG-VTON and SPG-VTON, respectively. Then, they were enlarged to 3.5 times the original size.}
  \label{fig:compare}
\end{figure*}

\begin{table*}[t]
\centering
\caption{Quantitative results of different variants of SPG-VTON on the MPV dataset. We highlight the \textbf{best} and the \underline{\textbf{second-best}} performances. $\uparrow$ denotes that higher scores are better. The variant ours w/ $ S_{t}$ represents that our method uses the ground-truth semantic map $ S_{t}$ of the test images to replace the predicted semantic map $\hat{S_{t}}$.}
\label{table:ab}
\vspace{-.5em} \scriptsize
\setlength{\tabcolsep}{12.5pt}
\begin{tabular}{l|c|c|c|c|c|c}
	\toprule
       \multirow{2}{*}{ Method} & \multirow{2}{*}{SSIM $\uparrow$} & \multirow{2}{*}{IS $\uparrow$ }& \multicolumn{2}{c|}{ Mask-SSIM $\uparrow$}& \multicolumn{2}{c}{Mask-IS $\uparrow$}\\
    \cmidrule{4-7} 
        & & & Face  &  Clothes & Face  &  Clothes \\
    \midrule
        Real data& 1.000&3.391 $\pm$ 0.024 & 1.000& 1.000& 2.255 $\pm$ 0.030 & 5.094 $ \pm$ 0.084  \\
     \midrule  
        Ours (\emph{full}) & 0.752 $\pm$ 0.084 & 3.243 $\pm$ 0.127 & 0.801 $\pm$ 0.127 & 0.872 $\pm$ 0.076 & 1.982 $\pm$ 0.024 & 5.057 $\pm$ 0.064 \\
      \midrule   
        w/o \emph{E2E} & 0.747 $\pm$ 0.087 & 2.968 $\pm$ 0.031 & 0.795 $\pm$ 0.130 & 0.869 $\pm$ 0.081 & 1.976 $\pm$ 0.013 &  4.854 $\pm$ 0.019\\
        w/o \emph{Cycle} & 0.739 $\pm$ 0.086 & 2.745 $\pm$ 0.036  & 0.794 $\pm$ 0.127 & 0.867 $\pm$ 0.080 & 1.801 $ \pm$ 0.016 & 4.841 $\pm$ 0.096\\ 
        w/o $ \hat{S_{t}}$ & 0.736 $\pm$ 0.088 & 2.647 $\pm$ 0.017 & 0.794 $\pm$ 0.129 & 0.861 $\pm$ 0.083 & 1.856 $ \pm$ 0.009  & 4.796 $\pm$ 0.027\\
        w/o $ L_{id}$& 0.732 $\pm$ 0.085 & 2.909 $\pm$ 0.069 & 0.780 $\pm$ 0.135 & 0.866 $\pm$ 0.080 & 1.969 $ \pm $ 0.045 & 4.601 $\pm$ 0.070 \\
        w/o $L^{l}_{adv}$ & 0.700 $ \pm$ 0.079 & \underline{\textbf{3.358 $\pm$ 0.022}} & 0.786 $ \pm$ 0.132 & 0.866 $\pm$ 0.080 &  1.830 $\pm$ 0.017 & 4.546 $\pm$ 0.047 \\
        w/o $L_{perc}$ & 0.724 $ \pm$ 0.086 & \textbf{3.474 $\pm$ 0.052} & 0.798 $ \pm$ 0.129 & 0.866 $\pm$ 0.081 &  \underline{\textbf{2.056 $\pm$ 0.015}} & \textbf{5.373 $\pm$ 0.077} \\
        w/o $M^{c}_{p}$ & 0.747 $ \pm$ 0.086 & 2.845 $\pm$ 0.012 & 0.800 $ \pm$ 0.127 & 0.869 $\pm$ 0.079 & 1.866 $\pm$ 0.016 & 4.958 $\pm$ 0.047 \\
        w/o \emph{Noise} & \underline{\textbf{0.756 $ \pm$ 0.087}} & 3.254 $\pm$ 0.019 & \underline{\textbf{0.802 $ \pm$ 0.127}} & \underline{\textbf{0.874 $\pm$ 0.079}} & 1.989 $\pm$ 0.017 & 5.066 $\pm$ 0.057 \\
        \midrule  
        w/ $S_{t}$  & \textbf{0.788 $\pm$ 0.080} & 3.165 $\pm$ 0.028 & \textbf{0.893 $\pm$ 0.081} & \textbf{0.923 $\pm$ 0.063} & \textbf{2.070 $\pm$ 0.013} & \underline{\textbf{5.186 $\pm$ 0.071}}\\ 
    \bottomrule
\end{tabular}

\end{table*}

\textbf{Setting.} In this work, we use the Adam optimizer~\cite{kingma2014adam} to optimize generators and discriminators in SPG-VTON and set the initial learning rate to 0.0002, weight decay to 0.0005, and exponential decay rates $(\beta_{1},\beta_{2})=(0,0.999)$. For training, we set hyperparameters $ \lambda_{i} = 10,
~(i=1,2,4,5)$, and $\lambda_{3} = 20$. Additionally, we set the batch size of semantic map prediction (first process of SPM, generator $G_{1}$) to 16, and the batch size of subsequent steps (generators $G_{2}$ and $G_{3}$, geometric matching modules GMM 1 and GMM 2) to 8, and the batch size for joint training is set to 8. In end-to-end training, we train the semantic map prediction network for $70k$, and then with the fixed pretrained semantic map prediction network, the subsequent network is trained for $70k$. Finally, we jointly train the whole model for $100k$.



\subsection{Quantitative Results}
\label{4.6}
We compare the proposed method with several start-of-the-art virtual try-on methods, including VITON~\cite{han2018viton}, CP-VTON~\cite{wang2018toward}, and MG-VTON~\cite{dong2019towards}. 
VITON and CP-VTON adopt a coarse-to-fine strategy to tackle the virtual try-on task of the single pose, and neither of these two methods includes the change of human pose. To make a fair comparison, we first enrich the input of VITON and CP-VTON by adding the target pose.
We report the quantitative results based on the SSIM and IS metrics (higher scores are better) to evaluate the realism of the synthesized virtual try-on images. As shown in TABLE \ref{tab:ssim_is}, our method achieves the maximum SSIM scores, and the maximum IS score on the MPV dataset. In addition, our method also obtains the highest IS scores on the DeepFashion dataset.
The results verify the effectiveness of the proposed method on generating high-fidelity virtual try-on images.

\begin{table*}[t]
\centering
\caption{Quantitative results of the effectiveness of five hyperparameters. We highlight the \textbf{best} performances. $\uparrow$ denotes that higher scores are better.}
\label{table:hyper}
\vspace{-.5em}\scriptsize
\setlength{\tabcolsep}{12.5pt}
\begin{tabular}{l|c|c|c|c|c|c}
	\toprule
       \multirow{2}{*}{ $\lambda$} & \multirow{2}{*}{SSIM $\uparrow$} & \multirow{2}{*}{IS $\uparrow$ }& \multicolumn{2}{c|}{ Mask-SSIM $\uparrow$}& \multicolumn{2}{c}{Mask-IS $\uparrow$}\\
    \cmidrule{4-7} 
        & & & Face  &  Clothes & Face  &  Clothes \\
    \midrule       
        $\lambda_{1}=0$  & 0.714 $\pm$ 0.078 & 2.983 $\pm$ 0.042 & 0.781 $\pm$ 0.131 & 0.852 $\pm$ 0.086 & 1.796 $\pm$ 0.006 & 4.916 $\pm$ 0.090 \\
        $\lambda_{2}=0$  & 0.554 $\pm$ 0.105 & \textbf{3.985 $\pm$ 0.052} & 0.784 $\pm$ 0.136 & 0.859 $\pm$ 0.081 & \textbf{2.121 $\pm$ 0.032} & 4.771 $\pm$ 0.048 \\
        $\lambda_{3}=0$ & 0.739 $\pm$ 0.085 & 2.581 $\pm$ 0.016 & 0.792 $\pm$ 0.130 & 0.869 $\pm$ 0.078 & 1.836 $\pm$ 0.018 & 4.151 $\pm$ 0.024 \\ 
        $\lambda_{4}=0$ & 0.420 $\pm$ 0.015 & 3.526 $\pm$ 0.059 &0.775 $\pm$ 0.136 & 0.857 $\pm$ 0.083 & 2.028 $\pm$ 0.013 & 4.858 $\pm$ 0.042 \\
        $\lambda_{5}=0$ & 0.732 $\pm$ 0.085 & 2.909 $\pm$ 0.069 & 0.780 $\pm$ 0.135 & 0.866 $\pm$ 0.080 & 1.969 $ \pm $ 0.045 & 4.601 $\pm$ 0.070 \\ 
    \midrule  
        Ours (\emph{full}) & \textbf{0.752 $\pm$ 0.084} & 3.243 $\pm$ 0.127 & \textbf{0.801 $\pm$ 0.127} & \textbf{0.872 $\pm$ 0.076} & 1.982 $\pm$ 0.024 & \textbf{5.057 $\pm$ 0.064} \\
    \bottomrule
\end{tabular}
\end{table*}

\textbf{Comparison with MG-VTON~\cite{dong2019towards}}. The multi-pose virtual try-on task aims to fit the desired clothes onto the target person image according to the source image and the given pose. Unlike the pose fixed virtual try-on issue, the multi-pose virtual try-on task is much more challenging to synthesize the whole try-on image, which does not preserve the original body parts. Since the existing methods (such as VITON~\cite{han2018viton} and CP-VTON~\cite{wang2018toward}) are based on the fixed pose and cannot find the spatial deformation relationship in pose change, directly applying these methods is inappropriate. MG-VTON is the first method to address the multi-pose virtual fitting issue. Hence, to further verify the effectiveness of this method, we conduct additional experiments to compare our method with MG-VTON on the MPV dataset. Consequently, we evaluate both the global metrics SSIM and IS and the local indicators Mask-SSIM and Mask-IS, which estimate the local (\ie, face region and clothes region) similarity and local realism between the generated image and the ground-truth image. Without the official code of MG-VTON, we calculate the results of MG-VTON from the try-on images provided by the original author. Note that these try-on images are generated from part of the test set. To make a fair comparison, we obtain our results from the same test images. Each test sample consists of a three-tuple, including source images, target poses, and desired clothes. The source image and target pose in each tuple in the test set have the same identity, so we calculate the Mask-SSIM value between the generated image and the image corresponding to the target pose. Specifically, we calculate Mask-SSIM of the facial area and the area unrelated to clothes between the generated image and the original image that provides the target pose.
As shown in TABLE \ref{tab:vsmg}, the proposed method achieves higher Mask-SSIM scores and Mask-IS scores than MG-VTON, suggesting that our method is better at fitting desired clothes onto the target human image.

\subsection{Qualitative Results}
We present visualized comparisons between MG-VTON and several variants of our method on the MPV dataset in Fig. \ref{fig:compare}. We observe that our method achieves higher-quality try-on results than MG-VTON. In particular, our method generates realistic and natural face regions while preserving the details of desired clothes. 

\begin{figure*}[t]
  \centering
  \includegraphics[width=\linewidth]{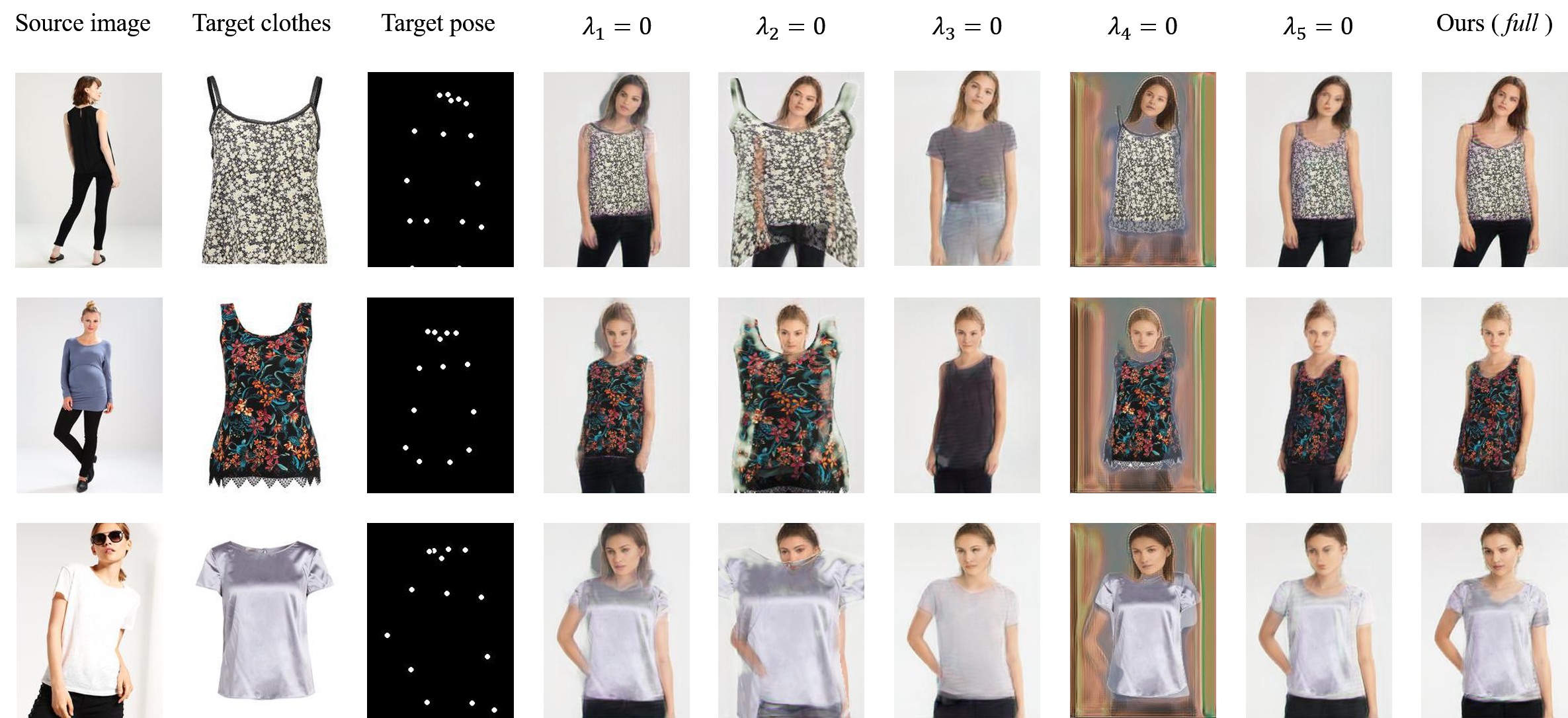}
  \vspace{-.5em}
  \caption{Qualitative visual comparison of the full model with the variants obtained by setting each of the five hyperparameters to zero.} 
  \label{fig:hyper}
\end{figure*}

\begin{figure}[t]
  \centering
  \includegraphics[width=\linewidth]{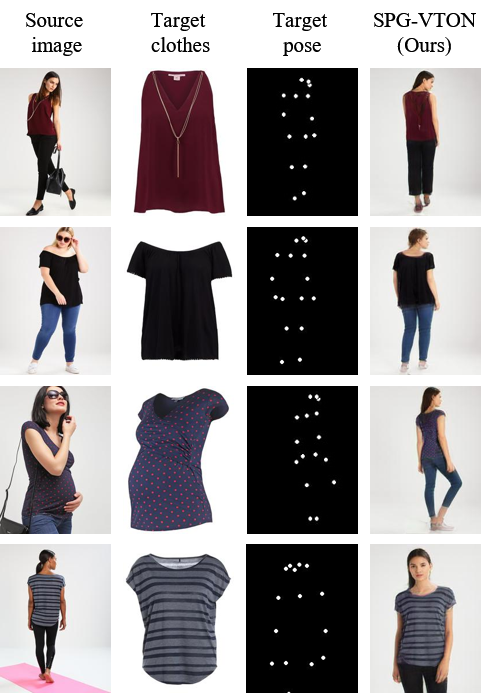}
  \vspace{-.5em}
  \caption{Examples of try-on results based on noisy images in the test set. For real-world applications, bags, sunglasses and Yoga mat are not desirable in the generated results. In this work, we view these objects as noise for the virtual try-on, and our method is still robust for such obstacles. The model has not seen the source inputs, which are all from the test set.
  }
  \label{fig:noise_vton}
\end{figure}

\subsection{Ablation Studies}
\label{4.7}
To study the effectiveness of each component in our method, we compare seven variants of SPG-VTON on the MPV dataset as follows: 
(1) w/o \emph{E2E}: our method without end-to-end training. Specifically, we train our method in a two-stage manner. We first separately train the semantic prediction network to obtain the preliminary semantic prediction map. Next, we use the semantic prediction map generated from the fixed pretrained semantic prediction network to guide the subsequent steps;
(2) w/o \emph{Cycle}: our method uses $\mathcal{L}_{1}$ loss between $C_{w}$ and $C_{t}$ to replace the conductible cycle consistency loss in the CWM. In this case, $ L_{cond} =\mathbb{E}[\| C_{w} - C_{t}\|_{1}]$;  
(3) w/o $ \hat{S_{t}}$: our method without the predicted semantic map $ \hat{S_{t}}$. In the training process, we use the predicted target body shape $ M^{b}_{p} $ to replace the predicted semantic map $\hat{S_{t}}$;
(4) w/o $ L_{id}$: our method removes the face identity loss. In this case, $L_{tsm} = \lambda_{4} (L^{t}_{recon} + L^{t}_{perc})+ L^{g}_{adv} + L^{l}_{adv}$;
(5) w/o $ L^{l}_{adv}$: our method without the local adversarial loss. In this case, $L_{tsm} = \lambda_{4} (L^{t}_{recon} + L^{t}_{perc})+ L^{g}_{adv} + \lambda_{5}L_{id}$;
(6) w/o $L_{perc}$: our method without the perceptual loss $L^{p}_{perc}$ and the perceptual loss $ L^{t}_{perc}$. In this case, $ L_{spm} = L^{s}_{adv} + \lambda_{1}L^{s}_{fl} + L^{s}_{recon} + L^{p}_{adv} +\lambda_{2}L^{p}_{recon}$, and
$L_{tsm} = \lambda_{4}L^{t}_{recon} + L^{g}_{adv} + L^{l}_{adv} +\lambda_{5}L_{id}$;
(7) w/o $M^{c}_{p}$: our method without the predicted clothing mask $M^{c}_{p}$. In the training process, we use the ground-truth clothing mask $M^{c}_{t}$ to replace the predicted clothing mask $M^{c}_{p}$. Accordingly, the TPS transformation parameters $\theta_{1}$ are calculated between the mask of desired clothes $ M^{C}$ and the ground-truth clothes mask $M^{c}_{t}$;
(8) w/o \emph{Noise}: we use filtered noise-free data to train the proposed method;
(9) w/ $ S_{t}$: our method uses the ground-truth semantic map $ S_{t}$ of the test images to replace the predicted semantic map $\hat{S_{t}}$. 

We show the qualitative results of several variants in Fig.~\ref{fig:compare}. We observe that without end-to-end training, our method can still produce plausible results. However, some artifacts existed near the hair regions and the neck regions in the generated images, suggesting that the end-to-end training manner could alleviate the impact of incorrect semantic maps on the generated results. We also find that without the cycle consistency constraint, the clothes regions in the generated images are deformed, which shows that the conductible cycle consistency loss could alleviate the mismatching between the desired clothes and the target person image. Additionally, we observe artifacts in the non-clothing regions of the generated results after replacing the predicted target semantic map with the predicted target body shape. This result indicates that without the guidance of the predicted semantic map, our method cannot accurately distinguish between regions of the human body. Moreover, we note that the facial regions in the generated results look less realistic after removing the global and local adversarial loss and the face identity loss, which suggests that introducing the global and local adversarial loss and the face identity loss can encourage the model to generate realistic and natural face regions. Additionally, we observe that removing the perceptual loss causes the model to generate blurred results with artifacts, suggesting that introducing the perceptual loss can synthesize sharper and clearer images. In addition, we find that when our method does not employ the predicted clothing mask, some artifacts appear in the generated image, and distortion exists in the clothing region. These results show that the process of target clothing mask prediction can encourage the model to accurately locate the clothing region and can alleviate the distortion between the desired clothes and the target pose.

We present the quantitative ablation study results of nine variants in TABLE \ref{table:ab}. We observe that the full model obtains higher SSIM scores and Mask-SSIM scores than seven different ablation methods except for the training set without noisy data and apply the ground-truth semantic map to
replace the predicted semantic map. We also note that without perceptual loss, the IS/Mask-IS scores are higher than those of the whole model.
This is due to the perceptual loss with deeper supervision (leveraging the activation of deeper layers), which focuses on semantic patterns. Therefore, the full model with the perceptual loss will improve the SSIM matching quality to the ground truth in terms of multiple scales but fail to hold better global matching IS. We surmise that this alternative is still open to the readers in considering whether to use the perceptual loss and that choices depend on the application, which emphasizes the pair matching performance or the global comparison with the whole dataset. The qualitative visual results with/without the perceptual loss are also provided in Fig.~\ref{fig:compare}. Additionally, we find that all indicators are decreased when our method does not implement the predicted clothing mask, suggesting that combining global semantic information with local semantic information could guide the proposed method to generate higher-quality fitting results.
More importantly, ablation studies also show that the conductible cycle consistency loss could match the shape of the deformed desired clothes with the target person image and maintain the characteristics of the desired clothes in the generated try-on image. We present the qualitative visual results with/without the cycle consistency constraint in Fig.~\ref{fig:compare}. Ablation studies also point out that using real semantic maps to replace predicted semantic maps at test time yields higher metric scores, suggesting that the semantic maps obtained by the semantic prediction networks still have a gap with real semantic maps. However, we find that all metrics increase with an end-to-end training manner. In particular, the IS/Mask-IS scores were superior. This result indicates that adopting an end-to-end training manner can enhance the accuracy of predicted semantic maps and lead the model to generate realistic fitting images.

\textbf{Impact of the introduced five hyperparameters. } To clarify the role of the five hyperparameters introduced in the objective function (\ie, Eq. \ref{loss1}), we first show the qualitative visual results for the full model with five variants in Fig.~\ref{fig:hyper}.
We observe that when $\lambda_{1}=0$, our method can still generate plausible fitting images, but artifacts appear. The main reason is that when $\lambda_{1}=0$, the focal loss is invalid. Therefore, the quality of the semantic map generated by SPM only maintains the rough body outline but cannot precisely locate the position of each body component. In addition, when $\lambda_{2}=0$, our method cannot generate reasonable fitting images. The main reason is that when $\lambda_{2}=0$, both
$L^{p}_{recon}$ and $L^{p}_{perc}$ are invalid, which leads to the inability of SPM to generate coarse results and predicted clothing masks, which are critical inputs for TSM and CWM, respectively. Meanwhile, when $\lambda_{3}=0$, we find that the details of the desired clothes are entirely lost in the generated images, which indicates that the conductible cycle consistency loss plays a vital role in the process of clothing deformation. Additionally, when $ \lambda_{4}=0$, we notice that even though the details of the facial and clothing regions in the generated image are well maintained, the entire quality of the generated image is poor due to the lack of global constraints. We also observe that when $\lambda_{5}=0$, the face identity loss is invalid, resulting in less realistic facial regions in the generated results.
Furthermore, we report the quantitative results of the effectiveness of five hyperparameters in TABLE \ref{table:hyper}. The full model obtains the highest SSIM/Mask-SSIM scores and the highest Mask-IS scores for the clothing regions in the generated images. Combining the results of qualitative and quantitative experiments, we consider it necessary to introduce these five hyperparameters.

\textbf{Impact of noisy images in the training set and the test set.} We conduct quantitative experiments to verify the impact of noisy data in the training set. As shown in Table \ref{table:ab}, we observe that all indicators (\ie, SSIM/Mask-SSIM, IS/Mask-IS) of the proposed method trained by noisy data are slightly lower than the variant trained with noise-free data. These results confirm that noisy training data cause the performance of our method to decrease, but the degree of decrease is still within an acceptable range. Notably, the test set still contains noisy data. Moreover, we provide qualitative visual results for virtual try-ons based on noisy images (\ie, source images with interference from attributes unrelated to the fitting task, such as bags, glasses, and complicated backgrounds). As shown in Fig.~\ref{fig:noise_vton}, we observe that the proposed method is robust to such training noise and shows good scalability to the unseen test images during inference.

\section{Conclusion}
In this paper, we propose a novel multi-pose virtual try-on framework (SPG-VTON) based on semantic prediction guidance, which focuses on producing photo-realistic try-on results while fitting the desired clothes onto an arbitrary pose of the same person. SPG-VTON consists of three submodules, including the semantic map prediction module, the clothing warping module, and the try-on synthesis module. On the one hand, we introduce a conductible cycle consistency loss that can alleviate the mismatching between the desired clothes and the target image. On the other hand, we also apply a face identity loss to make the face region of the final virtual try-on image look natural and to preserve the identity of the source image. Extensive qualitative and quantitative experiments demonstrate that the proposed method outperforms previous state-of-the-art methods and has good scalability to the training data noise as well as the unseen test images during inference. In the future, we will continue to explore the application of the proposed method to new fields, such as vehicle appearance design~\cite{zheng2020beyond} and language-based cloth generation~\cite{zheng2017dual}.



\ifCLASSOPTIONcaptionsoff
  \newpage
\fi



%

{\footnotesize
\bibliographystyle{IEEEtran}
\bibliography{egbib}
}


\end{document}